\documentclass[journal]{IEEEtran}
\ifCLASSINFOpdf
\else
\fi
\usepackage{amsmath}
\usepackage{graphicx}
\usepackage{algorithm}
\usepackage{algpseudocode}

\begin{document}
%
% paper title
% Titles are generally capitalized except for words such as a, an, and, as,
% at, but, by, for, in, nor, of, on, or, the, to and up, which are usually
% not capitalized unless they are the first or last word of the title.
% Linebreaks \\ can be used within to get better formatting as desired.
% Do not put math or special symbols in the title.
\title{Asynchronous Episodic Deep Deterministic Policy Gradient: Towards Continuous Control in Computationally Complex Environments}
%
%
% author names and IEEE memberships
% note positions of commas and nonbreaking spaces ( ~ ) LaTeX will not break
% a structure at a ~ so this keeps an author's name from being broken across
% two lines.
% use \thanks{} to gain access to the first footnote area
% a separate \thanks must be used for each paragraph as LaTeX2e's \thanks
% was not built to handle multiple paragraphs
%

\author{Zhizheng Zhang, Jiale Chen,
        Zhibo Chen,~\IEEEmembership{Senior Member,~IEEE,}
        and~Weiping~Li,~\IEEEmembership{Fellow,~IEEE}% <-this % stops a space
\thanks{Z. Zhang, J. Chen, Z. Chen and W. Li are with University of Science and Technology of China, Hefei, Anhui, 230026, China. (E-mail: chenzhibo@ustc.edu.cn)}}

% The paper headers
% \markboth{IEEE Transactions on Cybernetics.}%
\markboth{IEEE Transactions Submission.}%
{Shell \MakeLowercase{\textit{et al.}}: Bare Demo of IEEEtran.cls for IEEE Journals}
% \markboth{Journal of \LaTeX\ Class Files,~Vol.~14, No.~8, August~2015}%
% {Shell \MakeLowercase{\textit{et al.}}: Bare Demo of IEEEtran.cls for IEEE Journals}
% The only time the second header will appear is for the odd numbered pages
% after the title page when using the twoside option.
% 
% *** Note that you probably will NOT want to include the author's ***
% *** name in the headers of peer review papers.                   ***
% You can use \ifCLASSOPTIONpeerreview for conditional compilation here if
% you desire.

% If you want to put a publisher's ID mark on the page you can do it like
% this:
%\IEEEpubid{0000--0000/00\$00.00~\copyright~2015 IEEE}
% Remember, if you use this you must call \IEEEpubidadjcol in the second
% column for its text to clear the IEEEpubid mark.

% use for special paper notices
%\IEEEspecialpapernotice{(Invited Paper)}

% make the title area
\maketitle

% As a general rule, do not put math, special symbols or citations
% in the abstract or keywords.
\begin{abstract}
Deep Deterministic Policy Gradient (DDPG) has been proved to be a successful reinforcement learning (RL) algorithm for continuous control tasks. However, DDPG still suffers from data insufficiency and training inefficiency, especially in computationally complex environments. In this paper, we propose Asynchronous Episodic DDPG (AE-DDPG), as an expansion of DDPG, which can achieve more effective learning with less training time required. First, we design a modified scheme for data collection in an asynchronous fashion. Generally, for asynchronous RL algorithms, sample efficiency or/and training stability diminish as the degree of parallelism increases. We consider this problem from the perspectives of both data generation and data utilization. In detail, we re-design experience replay by introducing the idea of episodic control so that the agent can latch on good trajectories rapidly. In addition, we also inject a new type of noise in action space to enrich the exploration behaviors. Experiments demonstrate that our AE-DDPG achieves higher rewards and requires less time consuming than most popular RL algorithms in Learning to Run task which has a computationally complex environment. Not limited to the control tasks in computationally complex environments, AE-DDPG also achieves higher rewards and 2- to 4-fold improvement in sample efficiency on average compared to other variants of DDPG in MuJoCo environments. Furthermore, we verify the effectiveness of each proposed technique component through abundant ablation study.
\end{abstract}

% Note that keywords are not normally used for peerreview papers.

% For peer review papers, you can put extra information on the cover
% page as needed:
% \ifCLASSOPTIONpeerreview
% \begin{center} \bfseries EDICS Category: 3-BBND \end{center}
% \fi
%
% For peerreview papers, this IEEEtran command inserts a page break and
% creates the second title. It will be ignored for other modes.
\IEEEpeerreviewmaketitle

\section{Introduction}
% The very first letter is a 2 line initial drop letter followed
% by the rest of the first word in caps.
% 
% form to use if the first word consists of a single letter:
% \IEEEPARstart{A}{demo} file is ....
% 
% form to use if you need the single drop letter followed by
% normal text (unknown if ever used by the IEEE):
% \IEEEPARstart{A}{}demo file is ....
% 
% Some journals put the first two words in caps:
% \IEEEPARstart{T}{his demo} file is ....
% 
% Here we have the typical use of a "T" for an initial drop letter
% and "HIS" in caps to complete the first word.
\IEEEPARstart{D}{eep} neural networks have pushed further the envelope of reinforcement learning in a wide variety of domains, such as Atari games \cite{mnih2015human}, continuous systems control \cite{lillicrap2015continuous}, musculoskeletal models control for medical applications \cite{kidzinski2018learning}, etc. Deep reinforcement learning (Deep-RL) methods perform trail-and-error training through frequent interactions with the environments. Despite the impressive results, the problem of data insufficiency is still exposed seriously for Deep-RL in computationally complex environments, which leads to huge even intolerable time cost for training.

Data throughput and efficiency grossly dominate the performances of Deep-RL algorithms. Numerous distributed methods \cite{kuo1998parallel, mnih2016asynchronous, espeholt2018impala, horgan2018distributed, barth2018distributed, stooke2018accelerated} are proposed to solve this problem, the distributed implementations of which can be summarized into two categories: communicating gradients and communicating experiences. The popular distributed algorithm A3C \cite{mnih2016asynchronous} was proposed to improve data throughput by executing multiple agents in parallel and communicating gradients with respect to the parameters of the policy to a central parameter server. However, the distributed gradients calculation sacrifices the training stability, especially when the degree of parallelism increases or when the interaction become high-delayed in computationally complex environments. A feasible way capable of avoiding the training instability while increasing data throughput is to parallelize the processes of experience collection. The new scalable distributed Deep-RL agent IMPALA \cite{espeholt2018impala} adopts asynchronous experience collection for training a single agent on many tasks simultaneously. In IMPALA, multiple actors are used to interact with environments and communicate their trajectories with the learner responsible for policy updating. Although IMPALA has made great progress in solving multi-task problems, there are still some problems when performing parallel data collection on a single task in computationally complex environments. Similar with IMPALA, the Ape-X architecture in \cite{horgan2018distributed} and D4PG in \cite{barth2018distributed} extend the vanilla deep-learning-based frameworks to the distributed setting by involving a leaner network and multiple actor networks.

We argue that the data collection and the policy learning in an asynchronous framework are mutually influential with each other. Asynchronous data collection in off-policy Deep-RL methods facilitates exploring more potential strategies but increases the difficulty of distilling knowledge from the generated trajectories, which is also discussed in the early RL work \cite{baddeley2008reinforcement}. An intuitive reason for this issue is that it is more likely  to meet the mismatching problem between the speed of data collection and the speed of policy learning in asynchronous frameworks, which leads to a decline in the proportion of the the valuable samples for training and poor sample diversity. In this work, in addition to the asynchronous system, we focus on addressing two major challenges: 
\begin{itemize}
\item Sample imbalance. Data throughput is significantly increased due to asynchronous frameworks while small learning is still maintained to ensure training stability and avoid convergence to local optimal solutions. In this situation, parallelism of experience collection will aggravate sample imbalance where low-reward samples outnumber high-reward samples. 
\item Sample diversity. When asynchronously collecting experiences and training agents using off-policy methods on a single task, a lot of similar trajectories will be put into the same memory buffer for experience replay. Crucially, poor sample diversity will bring harmful affects to training efficiency. 
\end{itemize}

In this paper, with respect to continuous control in computationally complex environments, we propose Asynchronous Episodic DDPG (AE-DDPG) to address the aforementioned challenges. Unlike communicating gradients in A3C, the agent in AE-DDPG interacts with multiple stochastic environments simultaneously, which can achieve very high data throughput. To tackle the problem of sample imbalance, we employ the episodic control (EM) thinking \cite{blundell2016model}\cite{pritzel2017neural} in re-designing the experience replay of DDPG, which enables the agent to latch on high-reward policies rapidly. To the best of our knowledge, AE-DDPG is the first one that introduces episodic memory into Deep-RL methods for continuous problems. For the sake of improving sample diversity, we consider taking the power law signal with $(1/f)^2$ spectrum as noise injected in action space to enrich the agents' exploration behaviors.

We evaluate our proposed method on a realistic physiologically-based model control task, namely Learning to Run \cite{kidzinski2018learning}. Experimental results show that AE-DDPG outperforms not only the vanilla DDPG but also other popular RL methods in training efficiency and the resulting final policies. We won the 1st in the first round of NIPS 2017 Learning to Run Challenge by using this model. We also conduct experiments on other continuous tasks in MuJoCo environments to evaluate its generalization to other domains. Besides, we also verify the effectiveness of the technique components applied in AE-DDPG in our ablation study.

\section{Related Works}

\subsection{Asynchronous Methods for Deep-RL}
The early work \cite{kingma2014adam} had studied the convergence properties of Q-learning in the asynchronous settings. With the development of Deep-RL, the popular asynchronous algorithm A3C \cite{mnih2016asynchronous} was proposed to reduce the training time. As described before, the key idea of A3C is to execute multiple workers and communicate gradients. However, training stability and sample efficiency are negatively affected as the number of workers increases. The newly proposed asynchronous architecture IMPALA \cite{espeholt2018impala}, Ape-X \cite{horgan2018distributed} and D4PG \cite{barth2018distributed} all decouple experience collection and policy updating by involving a learner and multiple actors, in which actors need to copy the parameters of the leaner for n steps interactions. Although policy-lag caused by copying parameters from the learner to workers is mitigated by V-trace algorithm in \cite{espeholt2018impala}, the continuity of temporally correlated exploration in action space \cite{lillicrap2015continuous} will be affected harmfully, especially when training agents on a continuous control task with high-delayed interactions. Unlike communicating gradients in A3C and communicating experiences in IMPALA, Ape-X, D4PG, etc., we support for the setting of a single actor-critic pair and develop an asynchronous interactive mechanism to improve data throughput for the asynchronous implementation in continuous and computationally complex environments.

\subsection{Experience Replay}
Experience replay \cite{lin1992self} is a kind of technology that allows agents to reuse experience from the past. Prioritized experience replay \cite{schaul2015prioritized} weights the replay probabilities of experiences according to their measured temporal difference errors. But its additional run-time leads to diminished training efficiency as the number of trajectories increases. Hindsight experience replay \cite{andrychowicz2017hindsight} allows sample-efficient learning from the sparse and binary reward signals. In this paper, we aim to introduce the idea of episodic control to rapidly assimilate advanced knowledge from high-reward experiences and improve the diversity of actually sampled trajectories for experience replay.

\subsection{Episodic Control}
Episodic control is inspired by the functionality of hippocampus in the brain \cite{lengyel2008hippocampal}. The key idea of previous works on episodic control \cite{blundell2016model,pritzel2017neural} is to utilize highly rewarded experiences to help to recreate past successes in near-deterministic environments. Besides, episodic memory deep Q-networks  \cite{lin2018episodic} leverages episodic memory to regularize the learning target of deep Q-Networks rather than direct control. Note that episodic control in previous works requires table-based look-up in general. Therefore, it is mostly used to solve discrete problems in near-deterministic environments. Differently, we attempt to utilize episodic memory to encourage more effective experience replay in allusion to continuous control in complex stochastic environments.

\subsection{Noise for Exploration}
Noise for exploration in deep reinforcement learning mainly includes two categories: action space noise \cite{lillicrap2015continuous} and parameter space noise \cite{plappert2017parameter}. In terms of action space noise, uncorrelated Gaussian noise and noise based on the Ornstein-Uhlenbeck (OU) process are used mostly to teat the problem of exploration. In addition, parameter space noise \cite{plappert2017parameter} is also proposed to be an alternative acted on agents' parameters directly. In this paper, we introduce a new action space noise to alleviate the harmful affects of asynchronous experience collection on sample diversity.

\section{Background}
RL commonly models the trail-and-error learning procedures as the Markovian Decision Processes (MDP). At time \textit{t}, the agent observes the current state $s_{t}\in S$ of its interactive environment and chooses an action $a_{t}$ according to its policy $\mu_{\theta}(a|s_t), a\in A$. Then the environment returns the agent a scalar feedback signal $r_{t}\in R$ and translates to the next state according to the transition probability $P(s_{t+1}|s_{t}, a_t)$. The goal to find an optimal policy $\pi$ can be formulated as the mathematical problem of maximizing the expectation of cumulative discounted return $R_t = \sum_{t^{'}=t}^{\infty}\gamma^{t'-t}r_t$, where $\gamma \in [0,1)$ is the discount factor.

% \subsection{3.1 Deep Deterministic Policy Gradient (DDPG)}
DDPG \cite{lillicrap2015continuous} is an off-policy actor-critic algorithm \cite{grondman2012survey} proposed for continuous control with Deep-RL, which can be viewed as a successful modification to DPG algorithm \cite{silver2014deterministic}. DDPG consists of a neural network based policy function and a neural network based value function, which corresponds to the actor and the critic. We parameterize the actor $\mu$ and the critic $Q$ by $\theta^{\mu}$ and $\theta^{Q}$ respectively. Similarly, the target networks for actor and critic, parameterized by $\theta^{\mu'}$ and $\theta^{Q'}$ respectively, are introduced to alleviate the training instability in DDPG. We update the action distribution of the actor by applying policy gradient:
\begin{equation}\label{eq:1}
\nabla_{\theta^\mu}\approx \frac{1}{n}\sum_{i}\nabla_aQ(s_i,a|\theta^Q)\nabla_{\theta^\mu}\mu(s_i|\theta^\mu).
\end{equation}
We update critic by minimizing the loss:
\begin{equation}\label{eq:2}
\mathcal{L}(\theta^Q)=\frac{1}{n}\sum_{i}(y_i-Q(s_i,a_i|\theta_Q))^2.
\end{equation}
where
\begin{equation}\label{eq:3}
y_i=r_i+\gamma Q'(s_{i+1},\mu'(s_{i+1}|\theta^{\mu'})).
\end{equation}
The target networks are updated by enabling them track the learned networks with $\tau\in(0,1]$:
\begin{equation}\label{eq:4}
\begin{aligned}
\theta^{Q'}\leftarrow\tau\theta^{Q}+(1-\tau)\theta^{Q'}, \\ 
\theta^{u'}\leftarrow\tau\theta^{u}+(1-\tau)\theta^{u'}.
\end{aligned}
\end{equation}
Previous works have modified the vanilla DDPG from different aspects. For example, MA-BDDPG  with multi-actor \cite{kalweit2017uncertainty} and Multi-DDPG with multi-critic \cite{yang2017multi} are 
representative variants that make use of bootstrapped models to improve the sample efficiency and training stability. Another notable modified version is the expansion introduced in robotics to solve mapless navigation problems \cite{tai2017virtual}. This variant separates the sample collecting process to another thread from the training thread in a direct way. However, it hasn't addressed the crucial issues we describe in the introduction.

\section{Methodology}
In this section, we first introduce the overall architecture of AE-DDPG, which is illustrated in Figure \ref{fig:framework}. We then elaborate the algorithm we design for experience replay in this asynchronous architecture. A new action space noise is introduced for exploration in the final.

\subsection{AE-DDPG Architecture}

The agent in AE-DDPG performs asynchronous experience collection and synchronous policy learning in training. We first introduce the asynchronous interaction to improve the data throughout in AE-DDPG. Different with the bootstrapped models, there is a single actor-critic pair in our proposed AE-DDPG. Asynchronous sample collection helps to collect more data for policy learning especially in computationally complex environments, wherein the interaction is commonly time consuming. We therefore enable the actor in AE-DDPG to interact with multiple stochastic environments simultaneously. To achieve it, we run multiple environment simulators in parallel threads. These environment simulators are initialized randomly on the same task. In this way, the actor in AE-DDPG interacts with these environments simulators in parallel for asynchronous sample collection. A notable difference with the parallel, accelerated RL framework in \cite{stooke2018accelerated} is that we needn't gather all individual observations into a batch for inference at each step. Hence, the random fluctuations and straggler effect described in \cite{stooke2018accelerated} can be alleviated effectively in our framework. However, we haven't taken account of better utilization of multiple CPU and GPU like \cite{stooke2018accelerated}. Instead, we aim to improve the sample efficiency and policy exploration for the distributed RL framework in this paper.

\begin{figure}
\centering
\includegraphics[scale=0.55]{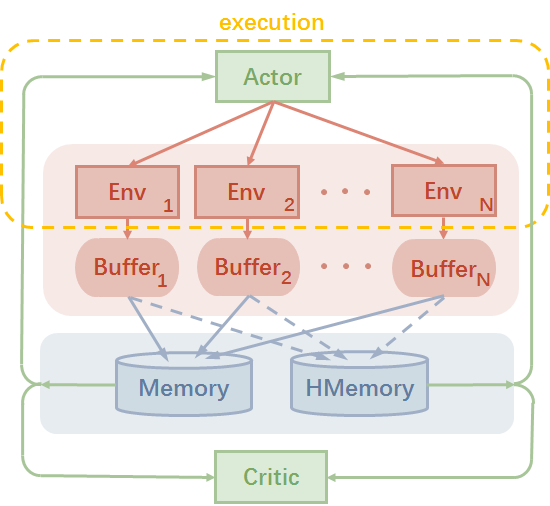}
\caption{Architecture of asynchronous episodic deep deterministic policy gradient (AE-DDPG).}
\label{fig:framework}
\end{figure}

We then introduce the memory buffers for experience replay in AE-DDPG. There are multiple experience cache buffers and two experience memory buffers in our proposed framework. As depicted in Figure \ref{fig:framework}, the trajectories generated by each interaction thread are cached into its individual cache buffer firstly. These buffers are thread-independent and they are not used for experience replay directly. Specially, we design two different memory buffers for the actual experience replay, they are ``Memory'' and ``HMemory''. Trajectories cached in the cache buffers will be put into the two memory buffers, namely ``Memory'' or/and ``HMemory'', according to the storing rule. Correspondingly, the trajectories stored in ``Memory'' or/and ``HMemory'' are sampled to be used for policy updating according to the sampling rule. Both the storing rule and the sampling rule are described in detail in our following section.

\subsection{Bio-inspired Episodic Experience Replay}
Asynchronous sample collection helps to improve data throughout especially when adopting RL algorithms in computationally complex environments, but this also leads to diminishing returns of sample efficiency as the degree of parallelism increases. It's easy to see that numerous similar interaction transitions are pushed into the memory when sharing experiences with a distributed RL framework. Therefore, we should balance the speed of data generation and the speed of data utilization to avoid worsening sample imbalance and improve the sample diversity. We aim to achieve this by proposing a novel experience replay.

Our design is inspired by the biological study on reward-motivated learning \cite{adcock2006reward}, in which the researchers use even-related FMRI to examine anticipatory mechanisms of reward-motivated memory formation. The result of 24-hr postscan suggests that subjects are significantly more likely to remember scenes that follows cues for high-value rather than low-value rewards. Additionally, a famous psychological research result, namely the so-called \textit{Peak-End Rule} \cite{do2008evaluations}, indicates that people are always sensitive to the peak returns or/and the end returns. 

Attempting to design a problem-solving RL algorithm with human-like efficiency and adaptability, we propose to decide which interaction trajectories should be attended more according to their cumulative reward at the end of one episode, briefly called ``episodic reward'' in our following description. To achieve this, we employ the idea of episodic control to improve experience replay. In detail, as described in the last section, we store the interaction trajectories into the thread-independent cache buffers and use two memory buffers for the actual experience replay. The two memory buffers have their respective functions. The module named ``Memory'' is similar with the memory buffer in vanilla DDPG, while the other ``HMemory'' is used for memorizing the highly rewarded trajectories. We use multiple threads to generate trajectories asynchronously and use a single thread to perform back-propagation on mini-batches of trajectories. The experience storing rule in our proposed bio-inspired episodic experience replay method is described in Algorithm \ref{alg:Storing}, and the rule of sampling mini-batches of trajectories for parameter updating is introduced in Algorithm \ref{alg:Sampling}.

\begin{algorithm}[thb]
  \caption{Storing Rule in Episodic Experience Replay}  
  \label{alg:Storing}
  $\mathcal{S}(s,a,r,s')$: a set of transitions in one episode. \\
  $(s_t,a_t,r_t,s_{t+1})$: one-step transition in trajectories. \\
  $\mathcal{N}_t$: action space noise in time $t$. \\
  $\mathcal{B}_i$: cache buffer for No.i interaction thread. \\
  $\mathcal{M}$: ``Memory'' depicted in Figure \ref{fig:framework}. \\
  $\mathcal{M}_H$: ``HMemory'' depicted in Figure \ref{fig:framework}. \\
  $R_e$: episodic reward (cumulative reward for one episode). \\
  $R_{max}^{EM}$: highest episodic reward in history. 
  \begin{algorithmic}[1]
  \For{each episode}
      \For{$t=1,2,\cdots,T$}
      \State Receive observation $o_t$ from environments.
      \State Get the state vector by $s_t=\phi(o_t)$.
      \State Execute action $a_t=\mu(s_t|\theta^\mu)+\mathcal{N}_t$.
      \State Observe reward $r_t$ and next state $s_{t+1}$.
      \State Store transition $(s_t,a_t,r_t,s_{t+1})$ in $\mathcal{B}_i$ and copy it to $\mathcal{M}$.
      \EndFor
      \State Pack all transitions into $\mathcal{S}(s,a,r,s')$.
      \State Calculate the episodic reward $R_e = \sum_{t=0}^{T_{end}}r_t$.
      \If{$R_e\geq R_{max}^{EM}$}
      \State Copy $\mathcal{S}(s,a,r,s')$ to $\mathcal{M}_H$.
      \EndIf    
      \State Update $R_{max}^{EM}\leftarrow max\{R_e,R_{max}^{EM}\}$.
  \EndFor
  \end{algorithmic}
\end{algorithm}

\begin{algorithm}[thb]  
  \caption{Sampling Rule in Episodic Experience Replay}  
  \label{alg:Sampling}
  $N_s$: number of transitions in a mini-batch. \\ 
  $\rho$: hyper-parameter, probability of sampling from ``HMemory''. \\
  (Definitions of other symbols are the same as Algorithm \ref{alg:Storing}.)
  \begin{algorithmic}[1]
  \For{each sampling}
      \For{$n=1,2,\cdots,N_s$}
      \State Generate a random number $m\in U(0,1)$.
      \If{$m\leq\rho$}
      	\State Sample transition $(s,a,r,s')$ from $\mathcal{M}_H$.
      \Else
      	\State Sample transition $(s,a,r,s')$ from $\mathcal{M}$.
      \EndIf
      \State Group all sampled transitions into a mini-batch.
      \EndFor
  \EndFor
  \end{algorithmic}
\end{algorithm}

Note that both of the two memory buffers are FIFO (First In First Out) buffers with limited memory space. The size of ``HMemory'' should be set smaller than the size of ``Memory''. We find two issues when training with the proposed experience replay: (1) highly rewarded trajectories are sampled more frequently; (2) low-reward trajectories in ``HMemory'' are easy to be dequeued relatively. These are consistent with our intuition that people are sensible to their best experiences, and they always tend to memorize the best experiences and learn from them.

\subsection{Random Walk Noise for Exploration}
Commonly, we explore the potential policy strategies in RL by adding a perturbation for the model parameters or the output actions of the RL agent. The latter one is called ``action space noise''. Here we denote the action space noise by $\mathcal{N}_t$ and formulate its usage as below:
\begin{equation}\label{eq:5}
\widehat{\Pi_\theta}(S_t)=\Pi_\theta (S_t)+\mathcal{N}_t.
\end{equation}
Where $S_t$ denotes the current state of the environment at time $t$, and $\Pi_\theta (\cdot)$ denotes the policy function. We obtain the practical action $\widehat{\Pi_\theta}(S_t)$, namely the control signal, by adding a random signal $\mathcal{N}_t$ to the output of actor network $\Pi_\theta(S_t)$. The noise only used for the exploration in the training stage, while $\Pi_\theta(S_t)$ are takes as the control signal directly in the testing stage.

Intuitively, the noise for exploration in continuous control problems should be not only temporally correlated but also instance uncorrelated. We need temporally correlated noise signals for exploration with respect to this type of problems because the executed actions are continuous in time. Thus, the temporally correlated signals benefit exploring more potential actions corresponding to better continuous control policy. The instance uncorrelated signals refer to that one of sampled sequence is uncorrelated with the sequence generated by another sampling process. Thus, the instance uncorrelated noise signals contribute to avoiding repeated and redundant exploration behaviors.

Here we propose to inject one of power law noise into the action space for policy exploration. Power law noise \cite{timmer1995generating} refers to a set of signals that exhibit a $(1/f)^\beta$ spectrum. Theoretically, the process of power low noise with $\beta=2$ is substantially equal to a rand-walk, which meets the two requirements we described above. We therefore adopt power law noise with $(1/f)^2$ spectrum to address the problem of exploration in continuous control tasks. In the following description, we will give more detailed explanations and introduce how we can generate this type of signals. 

We can obtain power law noise with $(1/f)^2$ spectrum simply by a first order filtering of white noise. Mathematically, we consider power law noise with $(1/f)^2$ spectrum as the realization of a random process $y(t)$ and take white noise as the realization of another random process $x(t)$. Because the $(1/f)^2$ spectrum power law signal can be obtained by filtering white noise, the relation of their spectral density can be formulated as:
\begin{equation}\label{eq:6}
S_y (\omega)=|H(j\omega)|^2 S_x (\omega),
\end{equation}
where $S_x (\omega)$ and $S_y (\omega)$ represent the power spectral density of $x(t)$ and $y(t)$ respectively. The $H(j\omega)$ corresponds to the filter we need. Since the power spectral density of white noise $S_x (\omega)=1$, we need to design the filter in Equation (\ref{eq:6}) as:
\begin{equation}\label{eq:7}
|H(j\omega)|^2=1/\omega^2.
\end{equation}
An equivalent discrete z transform for Equation (\ref{eq:7}) is:
\begin{equation}\label{eq:8}
H(z^{-1} )=1/(1-z^{-1}).
\end{equation}
Therefore, we can represent the filtered signal as:
\begin{equation}\label{eq:9}
Y(z^{-1} )=H(z^{-1} )X(z^{-1} )=X(z^{-1} )/(1-z^{-1}).
\end{equation}
According to the result of inverse z transform, we can get the signal in time domain as the following equation:
\begin{equation}\label{eq:10}
y_t-y_{t-1}=x_t.
\end{equation}

The formulas from Equation (\ref{eq:6}) to Equation (\ref{eq:10}) show that the signals with $(1/f)^2$ spectrum can be obtained by filtering the shots that start with a standard Gaussian generator. The Equation (\ref{eq:10}) represents a one-state auto regressive (AR) filter which produces the current value of the noise on the basis of the previous value. This is why we call it random walk noise and why this type of noise is temporally correlated and instance uncorrelated. 

In terms of exploration, random walk noise is able to improve the efficiency of exploration by capturing the temporal correlation of actions. In addition, it also helps improving sample diversity by enriching exploration behaviors, because different instances generated by this process is uncorrelated.

\section{Experiments}
\subsection{Environments}
\begin{figure}[h]
\centering
\includegraphics[scale=0.45]{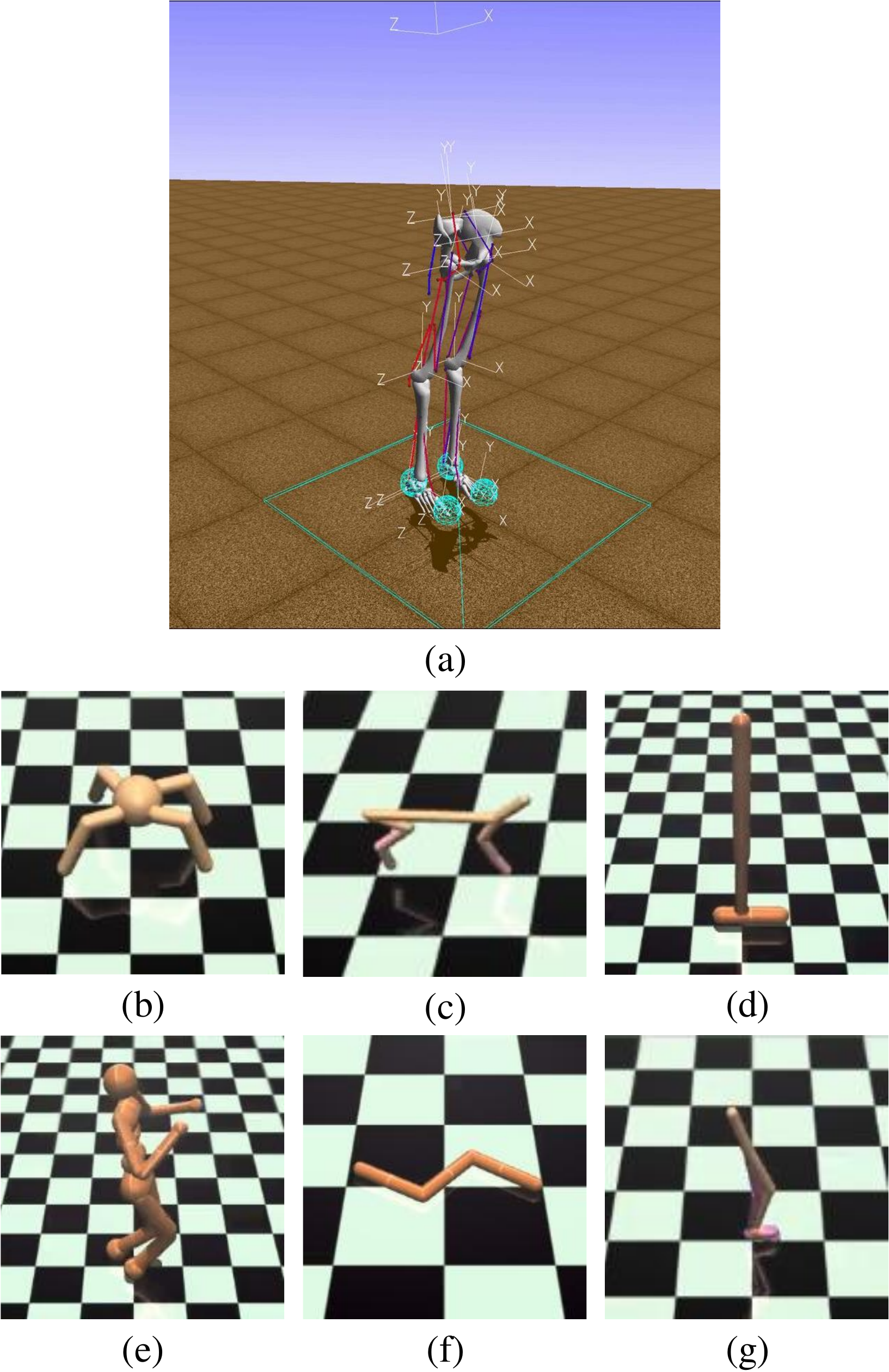}
\caption{Illustration of evaluation environments and tasks: (a) Musculoskeletal environment: Learning to Run; (b) MuJoCo: Ant-v2; (c) MuJoCo: Halfcheetah-v2. (d) MuJoCo: Hopper-v2; (e) MuJoCo: Humanoid-v2; (f) MuJoCo: Swimmer-v2; (g) MuJoCo: Walker2d-v2. }
\label{fig:env}
\end{figure}

We evaluate our proposed method, AE-DDPG, on a highly-simulated computationally complex environment and six continuous tasks in a standard benchmark platform. Furthermore, we conduct a series of experiments to verify the effectiveness of each proposed technique component within AE-DDPG. We first evaluate our proposed method AE-DDPG in a musculoskeletal environment with the task of controlling a highly-simulated human model to run like a human, namely \emph{Learning to Run} \cite{kidzinski2018learning}. We then compare AE-DDPG with other improved variants of DDPG on six continuous control tasks from OpenAI Gym \cite{brockman2016openai} simulated in MuJoCo \cite{todorov2012mujoco} to evaluate the generlization of AE-DDPG to other domains. Finally, we run ablation experiments on \emph{Learning to Run} environment. All of the simulation environments are illustrated in Figure \ref{fig:env}.

\begin{figure*}[htb]
\centering
\includegraphics[scale=0.38]{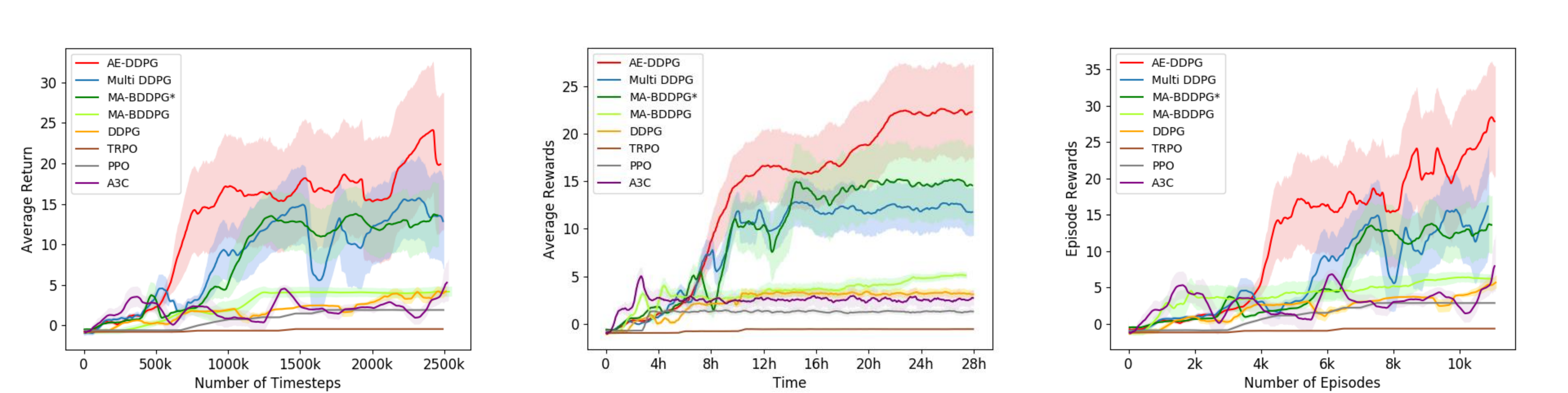}
\caption{Performance of our proposed AE-DDPG and other state-of-the-art RL algorithms on L2R task. To keep the same degree of parallelism for convincing comparison, we extent the original version of DA-BDDPG in \cite{kalweit2017uncertainty} by involving 16 actors to collect samples. And we denote the original version and expanded version by DA-BDDPG and DA-BDDPG$^*$ respectively.}
 \label{fig:final_result}
\end{figure*}

\subsubsection{Learning to Run (L2R) Environment}
The simulated environment of L2R task is implemented in OpenSim \cite{delp2007opensim} which is developed based on Simbody physics and biomechanics engine \cite{sherman2011simbody}. As shown in Figure \ref{fig:env}, a realistic physiologically-based human model is provided in this environment, which can achieve physically and physiologically accurate motion. Potential obstacles include external obstacles like stumbling blocks and the slippery floor, along with internal obstacles like materials weakness and motor noise. Besides, we can set different difficulty levels in L2R environment, which is corresponding to different number of randomly occurring stumbling blocks.

Given a 18-dimensional action vector corresponding to the excitations of simulated muscles, the environment engine will compute the physical force functions and return the status of the musculoskeletal model in the form of a 41-dimensional observation vector. The task of L2R is to control the provided human model to navigate a complex obstacle course as quickly as possible with the penalty of overusing ligaments taken into account.
\subsubsection{MuJoCo Environments} We use six continuous robotic control tasks from MuJoCo \cite{todorov2012mujoco} environments, running in a fast physical simulator, they are shown in Fig.\ref{fig:env}. The tasks of Ant-v2, HalfCheetah-v2, Hopper-v2, Humanoid-v2 and Walker2d-v2 are to control a four-legged creature model, a cheetah-like robot model, a two-dimensional one-legged robot, a humanoid robot and a two-dimensional bipedal robot respectively move forward as fast as possible. The task of Swimmer-v2 involves a 3-link swimming robot in a viscous fluid. In this task, we need to make it swim as fast as possible by actuating the two joints of the robotic model.

\begin{figure}
\centering
\includegraphics[scale=0.45]{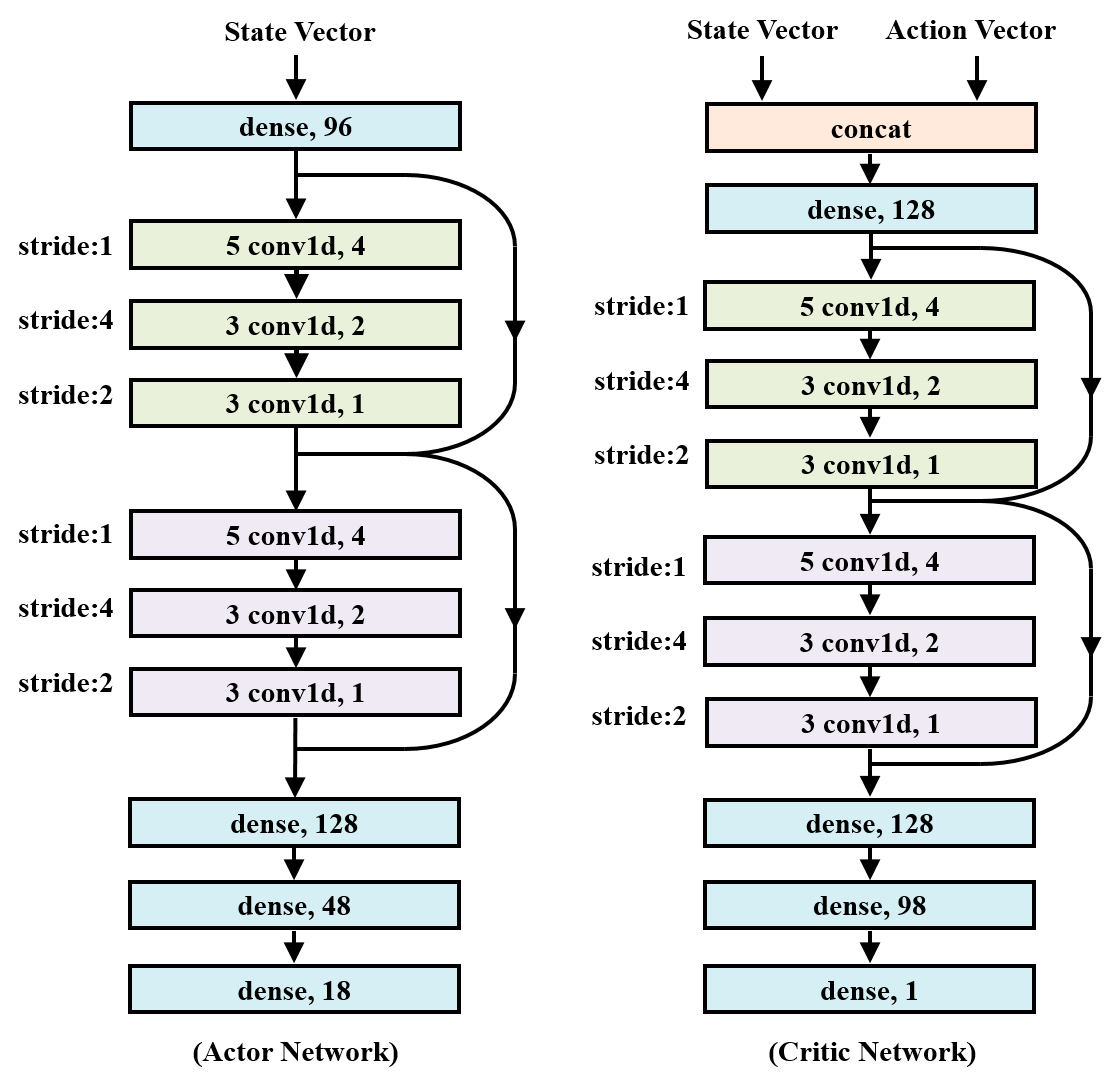}
\caption{The network architecture for L2R task. Each convolution layer is represented by its kernel size (for convolution), layer type and number of channels. LeakyReLU is applied in all layers except for the last layers of actor and critic. We use tanh activation for actor's last layer and linear activation for critic's last layer.}
 \label{fig:network}
\end{figure}

\subsection{Training Settings}
The difficulty level of L2R environment is set to be 2 for all of our experiments in this paper, which means that there are three stumbling blocks with random sizes and positions in each episode. To handle the high-dimensional observation vector and action vector, we specially design the network architecture depicted in Figure \ref{fig:network}. Adam \cite{kingma2014adam} is adopted to train the agent networks with a learning rate of $3e^{-4}$. We use mini-batch size $N=96$, discount factor $\gamma=0.99$, soft update rate $\tau=1e^{-3}$, and size of replay buffer $\mathcal{M}=10e^{6}, \mathcal{M}_H=5\times 10^{4}$ . Specially, we tune the probability $\rho$ of sampling from ``HMemory'' in $[0.05, 0.25]$ according to the number of interaction threads. 

In MuJoCo environments, we adopt fully connected networks with hidden sizes of (256, 256, 128) and (256, 128) to build the actor and critic respectively. And we use a learning rate of $1e^{-4}$ and a mini-batch size of 128. Other hyper-parameters keep the same settings of agents on L2R task.

\subsection{Results}
For convincing comparisons, the comparative models follow the settings of agents in AE-DDPG as possible. In this experimental setting, we try our best to reduce the variability of deep reinforcement learning caused by the potential factors discussed in \cite{henderson2017deep}. Therefore, when running the comparative models. we tune their own hyper-parameters, such as noise for exploration and experience replay, to enable them better performance. For each experimental case, we run 5 independent and repetitive experiments with different random seeds and report the best performance of them.

\begin{figure*}[ht]
\centering
\includegraphics[scale=0.52]{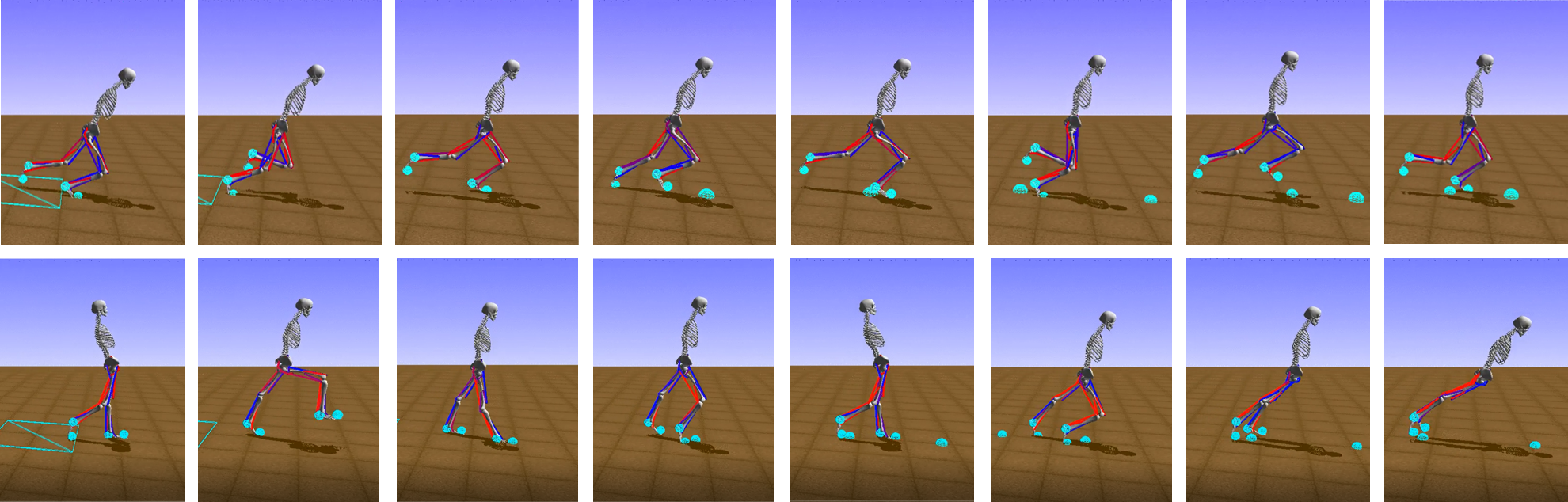}
\caption{The visualization results when training 10k episodes. Upper row: running postures of the agent trained by our AE-DDPG. Lower row: running postures of the agent trained by vanilla DDPG.}
 \label{fig:run}
\end{figure*}

\begin{figure*}
\centering
\includegraphics[scale=0.36]{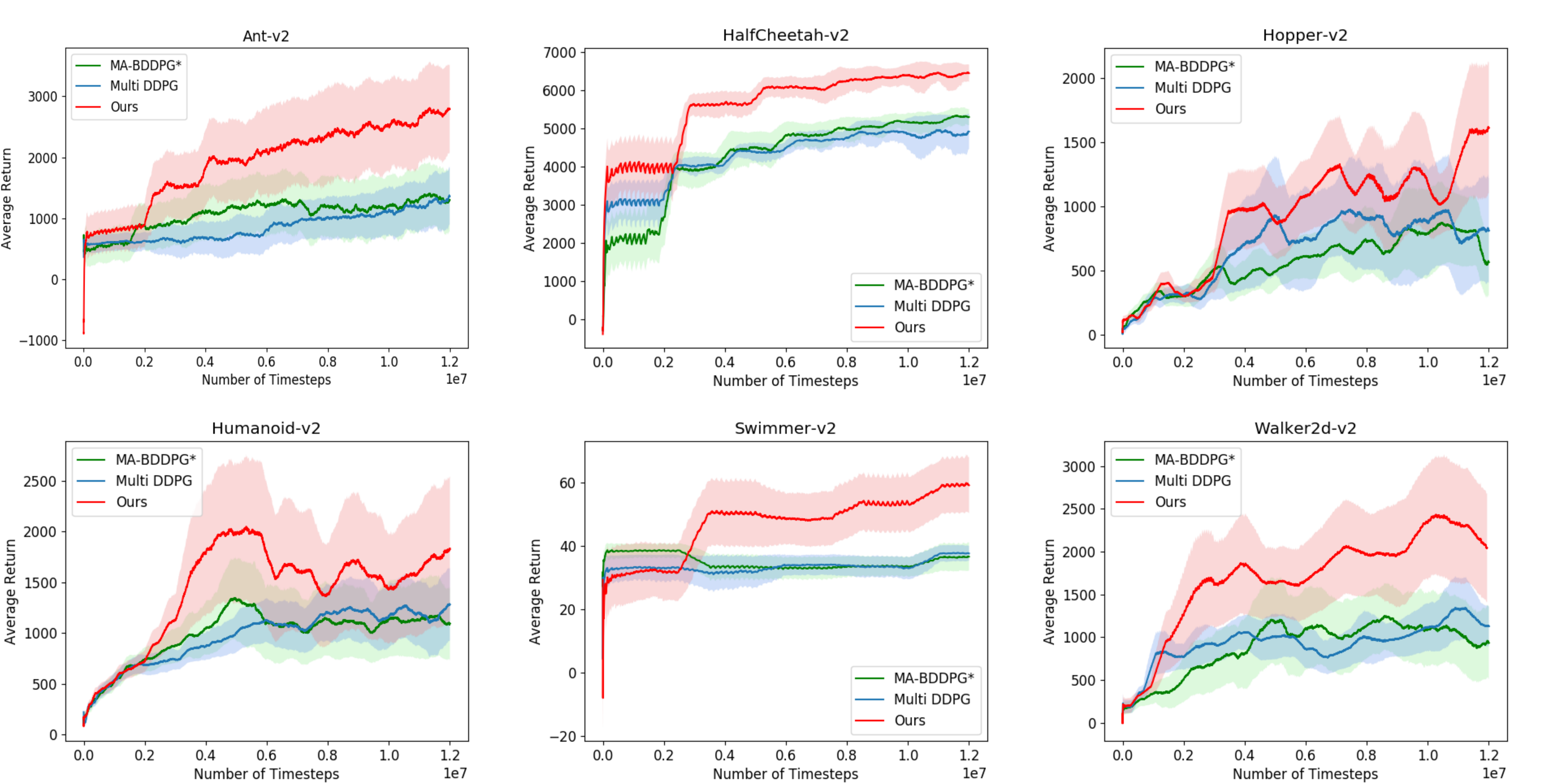}
\caption{Performance comparisons on six MuJoCo environments trained for 12 million timesteps, wherein one timestep equals one frame. The shaded region denotes the standard deviation over 5 random seeds.}
 \label{fig:mujoco}
\end{figure*}

\subsubsection{Evaluation in a Computationally Complex Environment} We compare our proposed method with both the state-of-the-art RL algorithms including three distributed variants of DDPG on L2R task. The MA-BDDPG \cite{kalweit2017uncertainty} and Multi-DDPG \cite{yang2017multi} used for comparison can be regarded as the distributed expansions of DDPG with multi-critic and multi-actor respectively. They tend to encourage data generation and estimate Q-values more accurately through introducing bootstrapped models. Different from both MA-BDDPG and Multi-DDPG, AE-DDPG is in defense of a paired actor-critic setting but has multiple environmental threads interacting with the actor asynchronously. For more convincing comparison, we further expand the vanilla version of MA-BDDPG by involving multiple actors to collect samples asynchronously, and we denote this expanded MA-BDDPG which has both multi-actor and multi-critic within one agent by ``MA-BDDPG$^*$'' in Fig. \ref{fig:final_result}. Note that we keep a single actor-critic pair but enable the actor to interact with multiple environment threads in our proposed AE-DDPG.

To evaluate our propose method in a fair setting, we have the actor in AE-DDPG interact with 16 stochastic L2R environments simultaneously and keep the same degree of parallelism in other algorithms (except vanilla DDPG and MA-BDDPG). Their best training performance across five repetitive experiments with different random seeds are reported in Figure \ref{fig:final_result}. The mean returns are represented by lines and std returns are represented by shaded areas.

In the left sub-figure of the Fig.\ref{fig:final_result}, given the same number of samples, AE-DDPG can achieve higher mean reward score. It indicates that AE-DDPG is more sample efficient than other algorithms for learning continuous control strategy in such a computationally complex environment. In the middle sub-figure, AE-DDPG can achieve higher mean reward score with requiring less time consuming than other algorithms when the training tends to be stable. We insist on that asynchronous sample collection benefits reducing time consuming especially in computationally complex environments, but it also leads to rapidly diminishing returns of sample efficiency due to the increasing sample imbalance and the decreasing of sample diversity as I mentioned in previous section. AE-DDPG shows strong ability in solving this problem by introducing bio-inspired episodic experience replay and random walk noise to encourage exploration and latch on the interaction trajectories rapidly. Implicitly, the right sub-figure shows better exploration ability for the potential strategies. 

A particularly notable issue is the comparison with A3C, where A3C is a little more effective in the beginning of the training but it fails in keeping this advantage in the follow-up learning. This might be caused by the mismatching between the speed of policy updating with gradient communication and the speed of data collection. We alleviate this problem by communicating experiences instead of communicating gradients. Here, IMPALA \cite{espeholt2018impala} is not taken into comparison since it is a set of scalable architectures designed for multi-tasks, and it doesn't address the issue of sample efficiency from the aspects of experience replay and noise.

We further visualize the running postures learned by different agents.
The human model trained by AE-DDPG is the closest one to a real adult runner. The model trained by the vanilla DDPG can move forward a few steps but it falls down soon (See Figure \ref{fig:run}). When training with Proximal Policy Optimization (PPO) \cite{schulman2017proximal}, the simulated human always keeps its two legs together and performs jump-like behaviors when runs forward. Trust Region Policy Optimization (TRPO) \cite{schulman2015trust} has weak effect on this task so that the human model trained by this algorithm has difficulty in keeping balance.

\begin{figure}[t]
\centering
\includegraphics[scale=0.5]{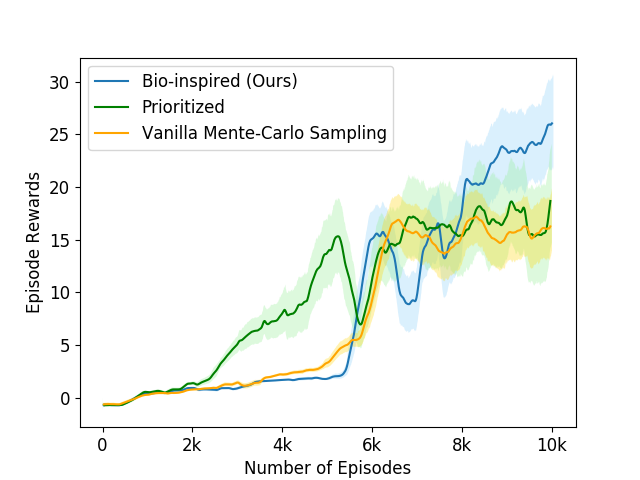}
\caption{Comparative experiments with different experience replay methods. Blue curve: bio-inspired experience replay (ours). Green curve: prioritized experience replay. Orange curve: original experience replay in vanilla DDPG.}
 \label{fig:memory}
\end{figure}

\subsubsection{Evaluation on a Standard Benchmark Platform}

To verify the generalization of our approach in other simple stochastic environments using a fast simulator, we choose six continuous MuJoCo environments from a standard benchmark platform OpenAI Gym as the evaluation tasks for this comparison experiment. Here, we compare a 16-thread AE-DDPG to the distributed DDPG-based variants, the modified MA-BDDPG and Multi-DDPG, in the introduced standard benchmark platform. 

In Fig. \ref{fig:mujoco}, we are glad to find that although the technologies in AE-DDPG are designed for continuous control in computationally complex environments, they are still effective and robust for the standard benchmark environments with fast simulators. Despite a single actor-critic pair within the agent, AE-DDPG has a high efficiency for data collection due to the asynchronous interactive mechanism. By enabling an actor interacting with multiple environments threads simultaneously, AE-DDPG can effectively avoid/alleviate the delay and the unstable interference caused by policy updating among the different actors or between the actors and the learners in the training. According to the performance comparisons show in Fig. \ref{fig:mujoco}, we insist on that our proposed experience relay and action space noise help the agent to explore the potential actions and distill useful information from them, which leads to high sample efficiency. The comparison results across six different tasks show that our proposed AE-DDPG with a single actor-critic pair has high sample efficiency exceeding the bootstrapped models.

\begin{figure}
\centering
\includegraphics[scale=0.5]{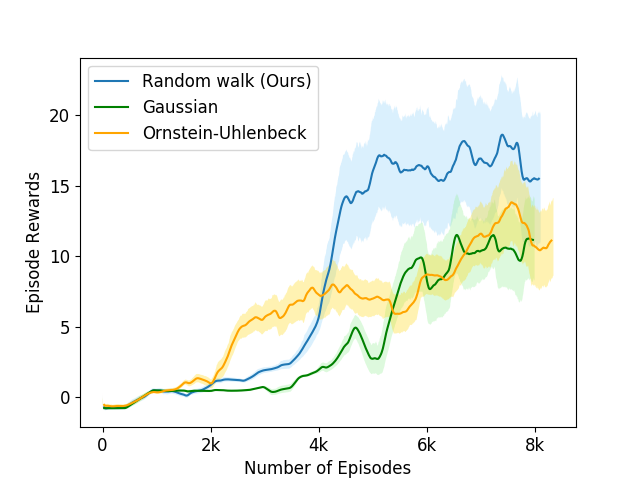}
\caption{Comparative experiments with different types of noise injected in action space. Blue curve: Random walk noise (ours). Green curve: noise sampled from Gaussian distribution ($\sigma=0.15$). Orange curve: noise generated by an Ornstein–Uhlenbeck process ($\sigma=0.2$).}
 \label{fig:noise}
\end{figure}

\subsubsection{Ablation Study for Episodic Experience Replay} 
To make clear the individual effects of our proposed experience replay, we conduct a series of experiments to compare it with prioritized experience replay \cite{schaul2015prioritized} and the original experience replay, that is vanilla monte-carlo sampling. With only different experience replay methods applied, other modules and settings in this set of comparative experiments remain the same with our above description.

The result depicted in Figure \ref{fig:memory} shows that episodic memory makes sense in alleviating the affects of sample imbalance by distilling important information from huge experiences collected from asynchronous interactions. The bio-inspired experience replay takes advantage of the insight from episodic control, which encourages the agent to pay more attention to highly rewarded trajectories. An interesting case here is that the prioritized experience replay shows the highest sample efficiency in the early stage of training, but the gain it brings declines gradually as experiences increase. This is because the significance of experience transitions is measured by TD errors in prioritized experience replay. However, the values of some actions might be overestimated and newly high-reward experiences are easy to be ignored, especially when using this method together with asynchronous experience collection. In general, the speed of experience generation mismatches the policy updating frequency more seriously in asynchronous frameworks. Because we generate more transitions in an asynchronous manner to find more potential high-reward actions, while we use small learning rate to keep the stability of gradient-based optimization. 

We also have an in-depth analysis for the mediocre performance of the episodic experience replay in the beginning stage. Storing rule of our proposed experience replay leads to short delay for the so-called ``HMemory'' buffer. Thus, the agent in AE-DDPG seems to take a conservative look at its potential success. 

\subsubsection{Ablation Study for Random Walk Noise}

To consider the possibility of developing our introduced noise as a plug-in technology, we further analysis its individual role. Therefore, we compare our proposed random walk noise with two popular noises injected in action space through the experiments on L2R task. One of them is sampled from Gaussian distribution while the other is generated by an Ornstein-Uhlenbeck (OU) process \cite{uhlenbeck1930theory}. We tune the parameters for each type of noise to reach their best performances for the convincing comparative results.

We can see that the RL agent using random walk noise achieves the highest mean reward score but with a relatively larger variance. This noise is proved to be successful in encouraging exploration behaviors. Its property of temporal correlation benefits finding effective actions in continuous space. In addition, the ``instance uncorrelated'' property of random walk noise helps to avoid repeating ineffective searches in the action space and substantially improve the sample diversity.

\section{Conclusion}
In this paper, an asynchronous actor-critic method AE-DDPG is proposed for developing a scalable and sample-efficient method to solve continuous control problems in computationally complex environments. Episodic control and power law noise with $(1/f)^2$ spectrum are successfully introduced in an asynchronous framework to help to remain even improve sample efficiency while increasing the data throughput. Experiments demonstrate that this modification of DDPG requires less training time and has higher learning efficiency in high-dimensional complex environments. It also shows the satisfactory generalization on other prevalent continuous tasks. We believe that the technique components inside AE-DDPG have the potential to be applied further in other Deep-RL algorithms in the future work.

\section*{Acknowledgment}

This work was supported in part by NSFC under Grant 61571413, 61632001,61390514.

% Can use something like this to put references on a page
% by themselves when using endfloat and the captionsoff option.
\ifCLASSOPTIONcaptionsoff
  \newpage
\fi

% trigger a \newpage just before the given reference
% number - used to balance the columns on the last page
% adjust value as needed - may need to be readjusted if
% the document is modified later
%\IEEEtriggeratref{8}
% The "triggered" command can be changed if desired:
%\IEEEtriggercmd{\enlargethispage{-5in}}

% references section

% can use a bibliography generated by BibTeX as a .bbl file
% BibTeX documentation can be easily obtained at:
% http://mirror.ctan.org/biblio/bibtex/contrib/doc/
% The IEEEtran BibTeX style support page is at:
% http://www.michaelshell.org/tex/ieeetran/bibtex/
%\bibliographystyle{IEEEtran}
% argument is your BibTeX string definitions and bibliography database(s)
%\bibliography{IEEEabrv,../bib/paper}
%
% <OR> manually copy in the resultant .bbl file
% set second argument of \begin to the number of references
% (used to reserve space for the reference number labels box)
% \begin{thebibliography}{1}

% \bibitem{IEEEhowto:kopka}
% H.~Kopka and P.~W. Daly, \emph{A Guide to \LaTeX}, 3rd~ed.\hskip 1em plus
%   0.5em minus 0.4em\relax Harlow, England: Addison-Wesley, 1999.

% \end{thebibliography}
\bibliographystyle{IEEEtran}
\bibliography{main}

\end{document}